\documentclass[runningheads]{llncs}
\usepackage{graphicx}
\usepackage{amsmath,amssymb} % define this before the line numbering.
\usepackage{color}
\usepackage{subfigure}
\usepackage{mathtools}
\usepackage{booktabs}
\usepackage{caption}
\usepackage{floatrow}
\usepackage{cite}
\usepackage[colorlinks,citecolor={green}]{hyperref}
\DeclarePairedDelimiter{\norm}{\lVert}{\rVert}
\begin{document}
\title{Deep Adversarial Attention Alignment for Unsupervised Domain Adaptation: \\
the Benefit of Target Expectation Maximization} 
% Replace with your title

\titlerunning{Deep Adversarial Attention Alignment for UDA}
% Replace with a meaningful short version of your title
%
\author{Guoliang Kang\inst{1} \and
Liang Zheng\inst{1,2} \and
Yan Yan\inst{1} \and 
%Zikun Liu\inst{3} \and 
Yi Yang\inst{1}}
%
%Please write out author names in full in the paper, i.e. full given and family names. 
%If any authors have names that can be parsed into FirstName LastName in multiple ways, please include the correct parsing, in a comment to the volume editors:
%\index{Lastnames, Firstnames}
%(Do not uncomment it, because you may introduce extra index items if you do that, we will use scripts for introducing index entries...)
\authorrunning{Guoliang Kang \emph{et al.}}
% Replace with shorter version of the author list. If there are more authors than fits a line, please use A. Author et al.
%

\institute{CAI, University of Technology Sydney \\
\email{\{Guoliang.Kang@student., Yan.Yan-3@student., Yi.Yang@\}uts.edu.au} \and 
Research School of Computer Science, Australian National University \\
\email{liangzheng06@gmail.com}
%\and
%Samsung Research Institute China - Beijing(SRC-B) \\
%\email{zikun.liu@samsung.com}
}
\maketitle              % typeset the header of the contribution
\begin{abstract}
In this paper, we make two contributions to unsupervised domain adaptation (UDA) using the convolutional neural network (CNN).
First, our approach transfers knowledge 
%tries to minimize the discrepancy across domains at the very start and 
in all
%the deep side of neural networks for all 
the convolutional layers through attention alignment. 
Most previous methods align high-level representations, 
\emph{e.g.,} activations of the fully connected (FC) layers. 
In these methods, however, the convolutional layers which underpin critical low-level domain knowledge cannot be updated directly towards reducing domain discrepancy. 
%disparities. 
%e.g., aligning the activations of fully-connected layers. 
%When aligning the FC layer,  the convolutional layers can be modified through gradient back-propagation,
%but not as expected.
%adjusting the parameters at the fully connected layers. 
%However, these methods may suffer from the decayed loss propagation after a few layers, especially when gradient explosion or vanishing takes place.
%While the modifications at the fully connected layer could be back propagated in principle, 
%it may decay 
%Our approach takes advantage of the natural image correspondence built by CycleGAN. 
Specifically, we assume that the discriminative regions in an image are relatively invariant to image style changes. Based on this assumption, we propose an attention alignment scheme on all the target convolutional layers to uncover the knowledge shared by the source domain. 
%Since the discriminative parts of an image are relatively invariant to image style changes, the proposed attention alignment is particularly suitable for robust knowledge adaptation.  
Second, we estimate the posterior label distribution of the unlabeled data for  target network training. 
Previous methods, which iteratively update the pseudo labels by the target network and refine the target network by the updated pseudo labels, are vulnerable to label estimation errors. 
Instead, our approach uses category distribution to calculate the cross-entropy loss for training,  thereby ameliorating the error accumulation of the estimated labels.  
The two contributions allow our approach to outperform the state-of-the-art methods by +2.6\% on the Office-31 dataset. 
%Notably, our approach yields +5.1\% improvement for the challenging \textbf{D} ${\rightarrow}$ \textbf{A} task.

\end{abstract}
\section{Introduction}
% the second row: the difference attention map for the same image with different models 
% the third row: the difference attention map for the real/fake images

\begin{figure}[t]
\par
%viewport=0 100 595 550
\includegraphics[scale=0.20]{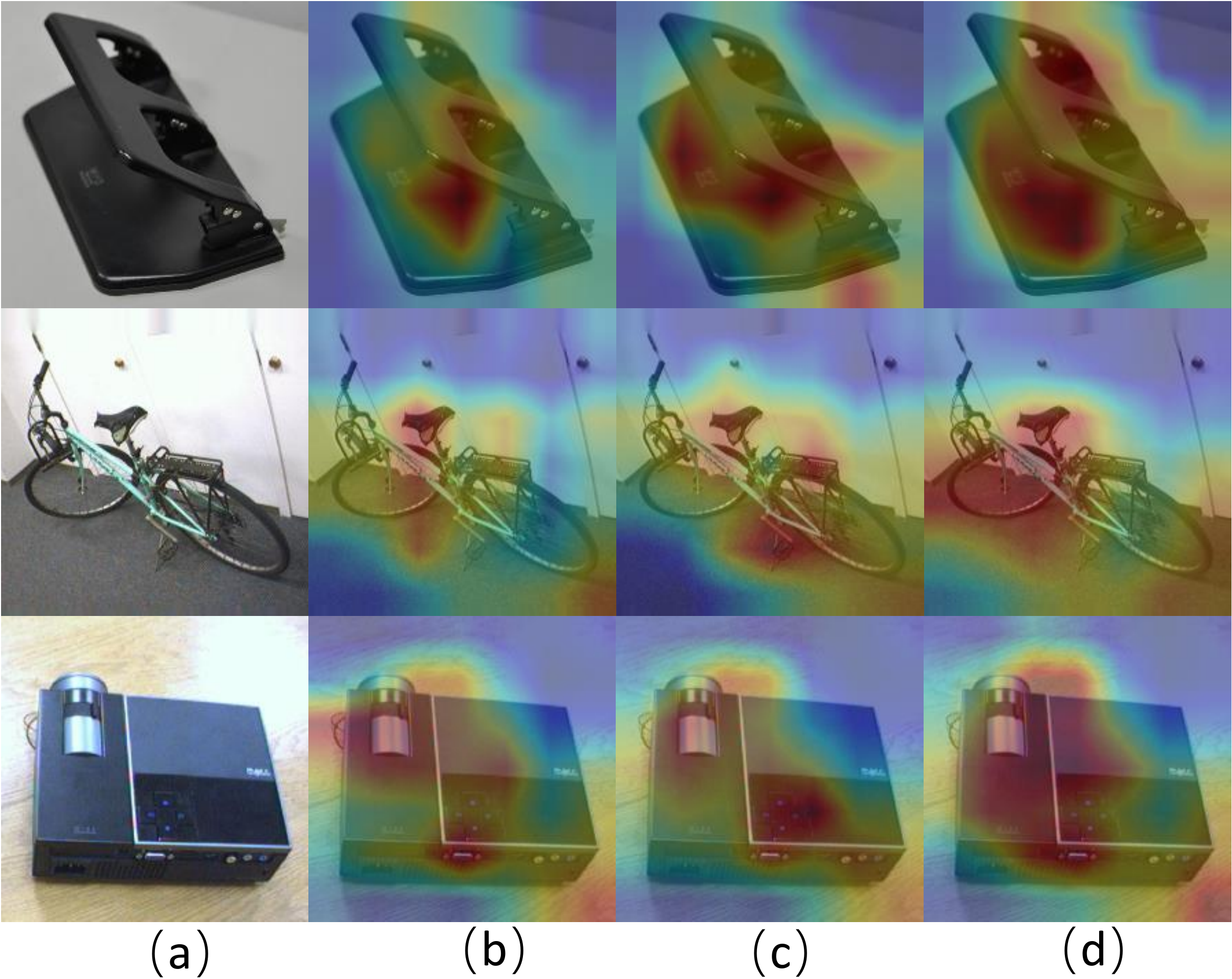}
\includegraphics[scale=0.20]{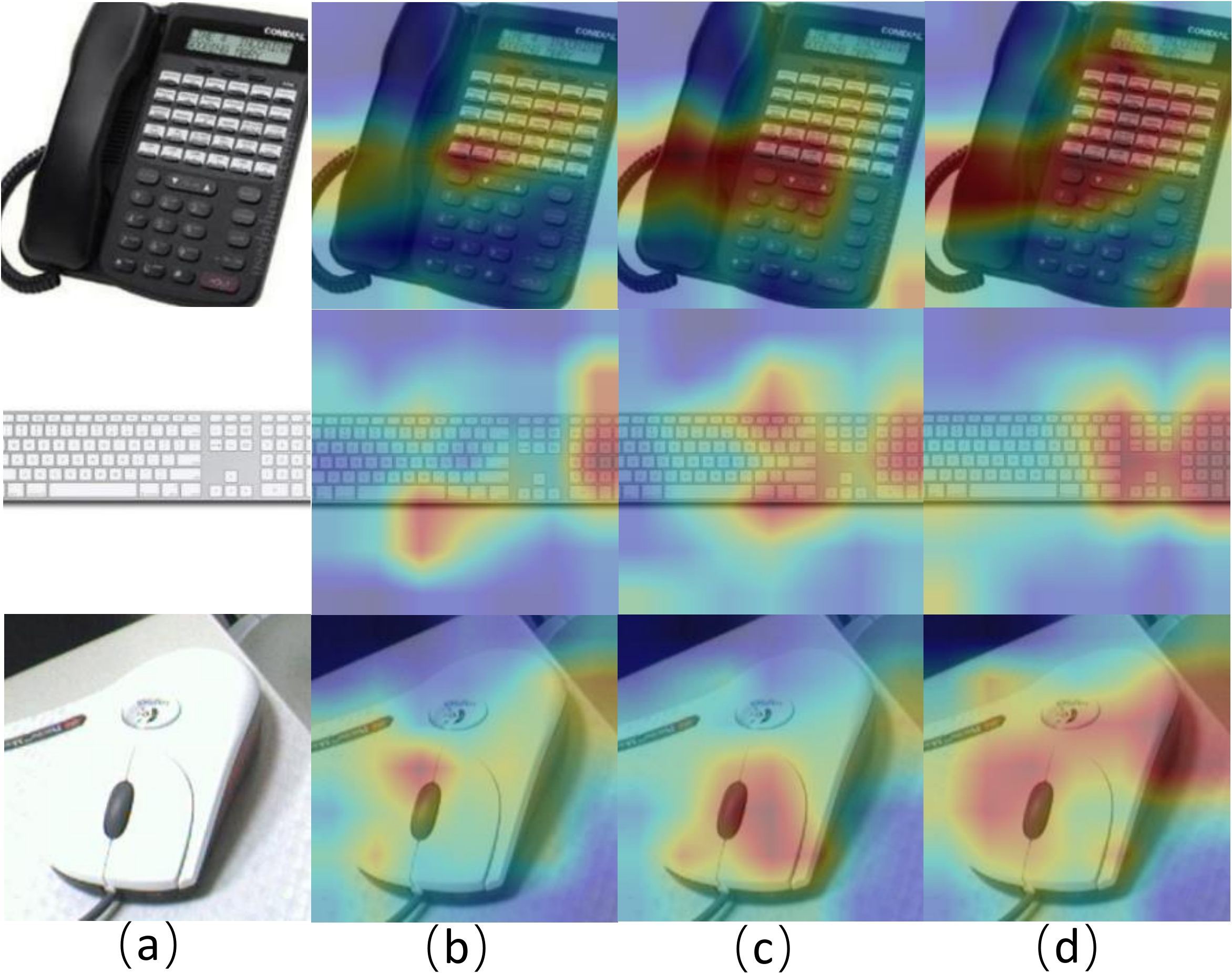}
\centering
\caption{\label{fig:attention-image}{
Attention visualization 
of the last convolutional layer of ResNet-50. 
%(\textbf{a}): The original target input image.
The original \textit{target} input images are illustrated in (\textbf{a}).
The corresponding attentions of 
the source network,
the target network trained on labeled target data,
and the target network adapted with adversarial attention alignment
are shown in (\textbf{b}), (\textbf{c}), and (\textbf{d}) respectively. 
%We observe that performing attention alignment between the target and the source network largely improves the attention mechanism on the target domain data. 
%Notably, the attention mechanism of the target network trained using 
%adversarial attention alignment is even better than that trained on its own domain labeled data, \emph{i.e.}, (\textbf{d}) is better than (\textbf{c}).
%Visualization of attention exhibited by CNNs fine-tuned on source and target domain (assume being labeled), respectively, for target images. By comparison,  CNN trained on the target domain pays more attention to the discriminative parts of the objects, while CNN trained on the source domain focuses more on the irrelevant background. This paper thus intends to perform attention alignment during domain adaptation. 
%which implies well-aligned attention cannot be reserved if we directly apply the model trained on one domain to another. 
%In each triplet, \textbf{Left}: a target image, 
%\textbf{Middle}: attention produced by CNN fine-tuned on the source domain,
%\textbf{Right}: attention produced by CNN fine-tuned on the target domain.
}} %\textbf{Left}}
\end{figure}

This paper focuses on unsupervised domain adaptation (UDA) for visual classification task. We aim to adapt the knowledge from a source network, trained by the source domain data, to the training of a target network, which will be used for making predications in the target domain. Note that in UDA the \emph{target domain is unlabeled}. The increasing popularity of UDA arises from the fact that the performance of a model trained on one domain may degenerate heavily on another when their underlying data distributions are different.

In the community of UDA, many  deep learning methods  %In recent years, the convolutional neural network (CNN) is widely adopted in the area of unsupervised domain adaptation.
attempt to minimize the discrepancy across domains on the top layers, such as the fully connected layers, of the neural network via explicitly imposing penalty terms 
\cite{tzeng2014deep,long2015learning,long2017deep,sun2016deep} or in an adversarial way \cite{ganin2015unsupervised,tzeng2017adversarial,tzeng2015simultaneous}. 
While the modifications at the fully connected layers can be back-propagated in principle, it may decay after a few layers, especially when gradient explosion 
or vanishing takes place. 
Consequently, the convolutional layers may be under-constrained. 
However, the domain discrepancy may emerge at the start from the convolutional layers, 
which makes any adjustment purely at the tail of the network less effective.
%If we could eliminate the domain discrepancy at the start
%from the convolutional layers, it would be more effective and efficient 
%to improve the adaptation performance.
%However, it remains unknown how to constrain the behavior of convolutional layers to enable them better adapted to the target domain data.

We investigate the domain discrepancy of the convolutional layers by visualizing their attention mechanisms.
In essence, the attention mechanism is emphasized as a key ingredient for CNN, suggested by a number of studies
\cite{simonyan2013deep,zeiler2014visualizing,zhou2016learning,selvaraju2017grad,zagoruyko2016paying,wei2017object,zhang2018adversarial}.
%Attention in CNN is similar to the mechanism that human being tends to focus on the salient parts in an image.
Zagoruyko \emph{et al.} \cite{zagoruyko2016paying} find that the model performance is highly correlated with the attention mechanism: a stronger model always owns better aligned attention than a weaker one. 
From Fig. \ref{fig:attention-image}, 
suppose we have networks trained on labeled data from source and target domains respectively,
we observe distinct attention patterns exhibited by the convolutional layers for the same target domain image.
The attention mechanism degenerates when directly applying the source network to the target domain data, 
which may exert negative influence on the classification performance.
%The fact that domain shift deteriorates the attention mechanism may exert negative influence on the classification system. 
Therefore, this paper expects the attention of the convolutional layers to be \emph{invariant to the domain shift}.

Based on the above discussions, 
%this paper proposes to align the attention of the target network 
this paper takes the domain discrepancy of the convolutional layers 
directly into account
by aligning the attention of the target network with the source network.
Our assumption is that no matter how domain varies, the discriminative parts of an image should be insensitive to the changes of image style.
%To measure the attention discrepancy across domains, 
Previous discrepancy measures (\emph{e.g.}, MMD \cite{long2015learning} and JMMD \cite{long2017deep})
which work effectively on high-level \textit{semantic} representations 
%distributions 
cannot be trivially transferred to measure the attention discrepancy 
of the convolutional layers 
where low-level \textit{structure} information is critical.
%In this procedure, the main obstacle to perform attention alignment is the lack of data correspondence. 
In this paper,
we propose using CycleGAN \cite{zhu2017unpaired} to build the data correspondence across domains,
\emph{i.e.}, translating the data 
from one domain to another without modifying its underlying content. 
Then, for the paired samples (\emph{e.g.} real source (or target) image and synthetic target (or source) image), 
we explicitly penalize the distances between attentions of the source and the target networks.
Additionally, we train our target network with real and synthetic data from both source and target domains. 
For source domain and its translated data, we impose the cross-entropy loss between the predictions and the ground-truth labels. 
For target domain and its translated source domain data, due to the lack of ground-truth labels, we make use of their underlying category distributions which provide insight into the target data. 
In a nutshell, we adopt the modified Expectation Maximization (EM) steps to maximize the likelihood of target domain images and update the model. Training iterations improve both the label posterior distribution estimation and the discriminative ability of the model. 
Our contributions are summarized below,

\begin{itemize}
\item We propose a deep attention alignment method which allows the target network to mimic the attention of the source network. Taking advantage of the pairing nature of CycleGAN, no additional supervision is needed. 
\item We propose using EM algorithm to exploit the unlabeled target data to update the network. 
Several modifications are made to stabilize  training and improve the adaptation performance.
\item Our method outperforms the state of art in all the six transfer tasks, achieving  +2.6\% improvement in average 
on the real-world domain adaptation dataset Office-31. 
\end{itemize}

  %We assume that the domain shift is largely due to the attention shift induced by the variation of domain data distributions. 
%We find that the attention mechanism should be invariant for the same task across domains. 
% We find that domain shift weakens the attention mechanism learned on the source domain.
% Based on CycleGAN pairing the samples across domains, we encourage the target network to mimic the attention of the teacher network by 
% explicitly penalizing the metric of distances of attention maps between the student and the teacher. 
% Experiment results demonstrate that through attention alignment, the performance of the target network on target domain data is 
% significantly improved, which proves that better attention mechanism will lead to higher accuracy.
%By building the data correspondence across domains through adversarial generative model, we successfully aligned the attention across different domains.

%%%%%%%%%%%%%%%%%%%%%%%%%%%%%%%%%
%------------------------------------------------------------------------
\section{Related Work} \label{related-work}
\textbf{Unsupervised domain adaptation.}
%(1. mainly the deep unsupervised domain adaptation methods; 2. what kind of knowledge should be transferred across domains; 3. our main contribution of indicating the attention mechanism of CNNs can be transferred across domains, which will consequently largely improves the model adaptation performance.)
% Deep domain confusion: Maximizing for domain invariance 2014
Various methods have been proposed for unsupervised domain adaptation 
\cite{tzeng2014deep,long2015learning,ganin2015unsupervised,long2017deep}.
%rozantsev2016beyond
Many works try to make the representations at the tail of neural networks invariant across domains. Tzeng \emph{et al.} \cite{tzeng2014deep} propose a kind of domain confusion loss to encourage the network to learn both semantically meaningful and domain invariant representations. 
Similarly, Long \emph{et al.} \cite{long2015learning} minimize the MMD distance of the fully-connected activations between source and target domain while sharing the convolutional features.
Ganin \emph{et al.} \cite{ganin2015unsupervised} enable the network to learn domain invariant representations in an adversarial way by adding a domain classifier and back-propagating inverse gradients.
JAN \cite{long2017deep} penalizes the JMMD over multiple fully-connected layers to minimize the domain discrepancy coming from both the data distribution and the label distribution.
Further, JAN-A \cite{long2017deep}, as a variant of JAN, trains the network in an adversarial way
with JMMD as the domain adversary.
DSN \cite{bousmalis2016domain} explicitly models domain-specific features to help improve networks'
ability to learn domain-invariant features.
%Haeusser \emph{et al.} \cite{haeusser2017associative} reinforce associations across domains directly in embedding space 
Associative domain adaptation (ADA) \cite{haeusser2017associative} reinforces associations across domains directly in embedding space to extract statistically domain-invariant and class discriminative features.
%Considering the source labels when minimizing the discrepancy across domains, 
Few works pay attention to the domain shift coming from the convolutional layers.
In this paper, we notice that the attention mechanism cannot be preserved when directly applying the model trained on the source domain to the target domain. To alleviate this problem, we constrain the training of convolutional layers by imposing the attention alignment penalty across domains. 

% Learning transferable features with deep adaptation networks 2015 *
% Deep transfer learning with joint adaptation networks 2016 *

% Deep coral: Correlation alignment for deep domain adaptation 2016
% Domain- adversarial training of neural networks 2016
% Adversarial Discriminative Domain Adaptation
% unsupervised pixel-level domain adaptation with generative adversarial networks

% “Simultaneous deep transfer across domains and tasks 2015
% Domain generalization for object recognition with multi-task autoencoders 2015
%
%-------------------------------------------------------------------------

\textbf{Attention of CNNs.}
% a. what is attention
%The study of the attention mechanism of computational models is inspired by human vision experience that human tends to focus their attention on the object of interest. Nowadays, attention is widely adopted in various applications, \emph{e.g.}, object recognition, visual captioning, and NLP. Here we focus on works \emph{w.r.t} understanding and representing visual attention mechanisms of CNNs.
% b. what is attention mechanism in deep neural network
% c. how to visualize and define the visual attention of a deep cnns
There exist many ways to define and visualize the attention mechanisms learned by CNNs.
%1） deconvolution network
Zeiler \& Fergus \cite{zeiler2014visualizing} project certain features back onto the image through a network called ``deconvnet'' which shares the same weights as the original feed-forward network.
Simonyan \emph{et al.} \cite{simonyan2013deep} propose using the gradient of the class score \emph{w.r.t} the input image to visualize the CNN.
Class activation maps (CAMs), proposed by \cite{zhou2016learning}, aim to visualize the class-discriminative image regions used by a CNN.
Grad-CAM \cite{selvaraju2017grad} combines gradient based attention method and CAM, enabling to obtain class-discriminative attention maps without modifying the original network structure as \cite{zhou2016learning}.

Zagoruyko \emph{et al.} \cite{zagoruyko2016paying} define attention as a set of spatial maps indicating which area the network focuses on to perform a certain task. 
The attention maps can also be defined \emph{w.r.t} various layers of the network so that they are able to capture both low-, mid-, and high-level representation information. 
They propose that attention mechanism should be a kind of knowledge transferred across different \textit{network architectures}.
Zaogruyko \emph{et al.} \cite{zagoruyko2016paying} align the attention across different architectures for exactly the same image during the training process and aim to transfer the 
knowledge from a large model to a smaller one.
Different to \cite{zagoruyko2016paying}, our method aligns the attention across different \textit{data domains} where images across domains 
are unpaired and aims to promote the model adaptation performance.

\textbf{Unpaired image-to-image translation.}
Unpaired image-to-image translation aims to 
train a model to map image
samples across domains, under the absence
of pairing information.
It can be realized through GAN to pair the real source (or target) and 
synthetic target (or source) images 
\cite{liu2016coupled,shrivastava2016learning,zhu2017unpaired,kim2017learning,liu2017unsupervised,bousmalis2017unsupervised,hoffman2017cycada,russo2017source}.
Generating synthetic images can be beneficial for various vision tasks 
\cite{luc2016semantic,zheng2017unlabeled,dong2018san,ding2015multilayer}.
In this paper, we concentrate on maximizing the utility of given paired real and synthetic samples. 
And we choose CycleGAN \cite{zhu2017unpaired} to perform such adversarial data pairing.
%A better GAN framework will promote the performance of our method, 
%however the discussions of which reach beyond the scope of this paper. 

% 1) CoGAN
% 2) DiscoGAN
% 3) CycleGAN
% 4) NIPS
%In this paper, we choose CycleGAN \cite{zhu2017unpaired} as our image-to-image translator to map images from source domain to target domain.

\begin{figure*}[t]
\begin{center}
\par
%\includegraphics[viewport=100 0 595 550,scale=0.35]{figure}
%width=0.99\textwidth
\includegraphics[width=0.9\textwidth]{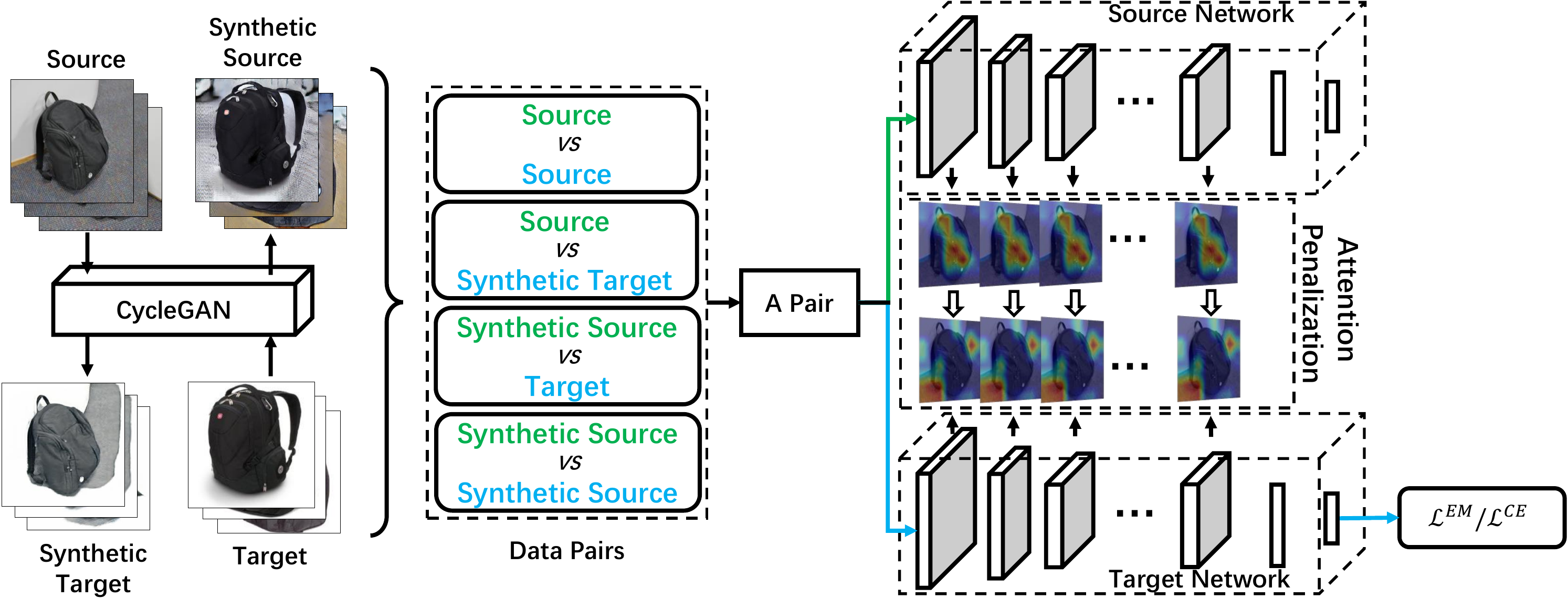}
\end{center}

\caption{\label{fig:schematic-diagram}{The framework of deep adversarial attention alignment. 
We train a source network and fix it. The source network guides the attention alignment of the target network.
The target network is trained with real and synthetic images from both domains. 
For labeled real source and synthetic target data, we update the network by computing the cross-entropy loss 
between the predictions and the ground-truth labels.
For unlabeled real target and synthetic source images, we maximize the likelihood of the data with EM steps.
The attention distance for a pair of images (as illustrated in the ``Data Pairs'' block) passing through 
the source network and the target network, respectively, is minimized.
}}
\end{figure*}

\section{Deep Adversarial Attention Alignment}
Our framework is illustrated in Fig. \ref{fig:schematic-diagram}. We train a source CNN which guides the attention alignment of the target CNN whose convolutional layers have the same architecture as the source network. 
The target CNN is trained with a mixture of real and synthetic images from both source and target domains. For source and synthetic target domain data, we have ground-truth 
labels and use them to train the target network with cross-entropy loss. 
On the other hand, for the target and synthetic source domain data, due to the lack of ground-truth labels, 
we optimize the target network through an EM algorithm.

\subsection{Adversarial Data Pairing}
%(1. introducing GAN and CycleGAN)
We use CycleGAN to translate the samples in the source domain ${S}$ to those in the target domain ${T}$, and vice versa. 
%Because the training data is unpaired, this process is called ``unpaired image-to-image translation".
The underlying assumption to obtain meaningful translation is that there exist some relationships between two domains.
For unsupervised domain adaptation, the objects of interest across domains belong to the same set of category. 
So it is possible to use CycleGAN to map the sample in the source domain to that in the 
target domain while maintaining the underlying object-of-interest.

% introduce CycleGAN
% what is GAN
% why CYcleGAN
The Generative Adversarial Network (GAN) aims to generate synthetic
images which are indistinguishable from real samples through an adversarial loss,
\begin{align}
\mathcal{L}^{GAN}(G^{ST}, D^T, X^S, X^T) &= \mathbb{E}_{x^T}[\log{D^T(x^T)}] %\nonumber \\
                       +\mathbb{E}_{x^S}[1-\log{D^T(G^{ST}(x^S))}],
\label{eq-gan-loss}
\end{align}
where $x^S$ and $x^T$ are sampled from source domain ${S}$ and target domain ${T}$, respectively.
The generator $G^{ST}$ mapping $X^S$ to $X^T$ strives to make its generated synthetic outputs $G^{ST}(x^S)$
indistinguishable from real target samples ${x^T}$ for the domain discriminator $D^T$.
% TODO: some explanations about the above formula

\begin{figure}[t]
\centering
\subfigure[]{
        \label{fig:real-source}
        \includegraphics[scale=0.20]{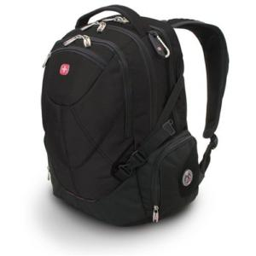}
        }
\subfigure[]{
        \label{fig:fake-target}
        \includegraphics[scale=0.20]{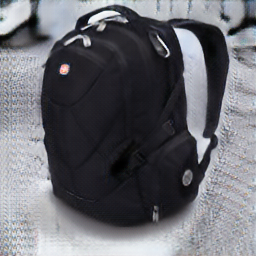}
        }
\subfigure[]{
        \label{fig:real-target}
        \includegraphics[scale=0.20]{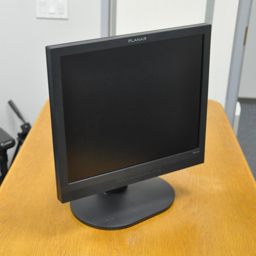}
        }%
\subfigure[]{
        \label{fig:fake-source}
        \includegraphics[scale=0.20]{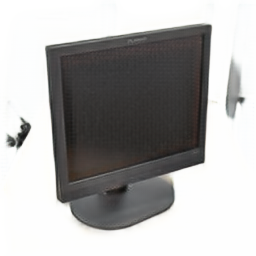}
        }%
%\end{center}
\caption{\label{fig:gan-images}{Paired data across domains using CycleGAN. (\textbf{a}) and (\textbf{c}): real images sampled from source and target domain, respectively. 
(\textbf{b}): a synthetic target image paired with (\textbf{a}) through $G^{ST}$. (\textbf{d}): a synthetic source image paired with a real target image (\textbf{c}) through $G^{TS}$.}}
\end{figure}

Because the training data across domains are unpaired, the translation from source domain to target domain is highly under-constrained.
CycleGAN couples the adversarial training of this mapping with its inverse one, \emph{i.e.} the mapping from $S$ to $T$
and that from ${T}$ to ${S}$ are learned concurrently. Moreover, it introduces a cycle consistency loss to regularize the training,
\begin{align}
\mathcal{L}^{cyc}(G^{ST}, G^{TS}) &= \mathbb{E}_{x^S}[\norm{G^{TS}(G^{ST}(x^S))-x^S}_1] 
                  +\mathbb{E}_{x^T}[\norm{G^{ST}(G^{TS}(x^T))-x^T}_1],
\label{eq-gan-cyc-loss}
\end{align}
Formally, the full objective for CycleGAN is, 
\begin{align}
\mathcal{L}^{cyc}(G, F, D_X, D_Y) &= \mathcal{L}^{GAN}(G^{ST}, D^T, X^S, X^T) 
                 +\mathcal{L}^{GAN}(G^{TS}, D^S, X^T, X^S) \nonumber \\
                 &+\lambda \mathcal{L}^{cyc}(G^{ST}, G^{TS}),
\label{eq-gan-full-loss}
\end{align}
where the constant $\lambda$ controls the strength of the cycle consistency loss.
Through CycleGAN, we are able to translate an image in the source domain to that in the target domain in the context of our visual domain adaptation 
tasks (Fig. \ref{fig:gan-images}).

As illustrated in Fig. \ref{fig:attention-image}, 
% no matter for the real ones and the synthetic ones generated from the corresponding source data through CycleGAN. 
the target model pays too much attention to the irrelevant background or less discriminative parts of the objects of interest. 
This attention misalignment will degenerate the model's performance. In this paper, we propose to use the style-translated images as natural image correspondences to guide the attention mechanism of the target model to mimic that of the source model, to be detailed in Section \ref{sec:attention}.
%source domain cannot be 
%directly transferred to the target data.
%In this section, we propose imposing the attention alignment penalty to reduce the extent of attention misalignment and enable the well-aligned attention mechanism leaned by the source domain data to be gradually transferred to the target domain data.

\subsection{Attention Alignment}\label{sec:attention}
Based on the paired images, we propose imposing the attention alignment penalty to reduce the discrepancy of attention maps across domains.
%Through CycleGAN, we have obtained synthetic target images with labels and 
%same contents as corresponding source images.
%For a pair of source and synthetic target images, the objects of interest 
%are the same. However, as shown in Fig. xx, the attentions are misaligned 
%with less discriminative parts in the image from target domain, although 
%they are quite well for the source domain images. So a question is, how 
%to transfer the attention mechanism from source to target domain, or how
%to enable the attention mechanism be adapted across domains?
Specifically, we represent  attention as a function of spatial maps \emph{w.r.t} each convolutional layer \cite{zagoruyko2016paying}. 
For the input $x$ of a CNN, let the corresponding feature maps \emph{w.r.t} layer $l$ be represented by $F_l(x)$. 
Then, the attention map $A_{l}(x)$ \emph{w.r.t} layer $l$ is defined as
\begin{equation}
A_{l}(x) = \sum_c|{F_{l,c}}(x)|^2,
\label{eq-atm-def}
\end{equation}
% TODO: feature maps or feature map
where $F_{l,c}(x)$ denotes the $c$-th channel of the feature maps. 
The operations in Eq. (\ref{eq-atm-def}) are all element-wise.
Alternative ways to represent the attention maps include $\sum_c|F_{l,c}|$, and $\max{|F_{l, c}|}$, \emph{etc}. 
We adopt Eq. (\ref{eq-atm-def}) to emphasize the salient parts of the feature maps. 

% TODO:
% two model: 
% one model trained on the source only data --> teacher
% one model trained on mixture of source and synthetic target data --> student
% what does teacher teach the student and why can it teach?
%We can train the model in fully supervised way on labeled source data and obtain well-aligned attention mechanism.
%It is expected that if this well-aligned attention mechanism could be transferred to the target domain, the adaptation performance 
%would be largely improved.
We propose using the source network to guide the attention alignment 
of the target network, as illustrated in Fig. \ref{fig:schematic-diagram}.
We penalize the distance between the vectorized attention maps between the source and the target networks 
to minimize their discrepancy. 
In order to make the attention mechanism invariant to the domain shift, 
we train the target network with a mixture of real and synthetic data from both source and target domains.

%For each sample $x$, let $\hat{A}$ and $A$ denote the attention maps obtained through the forward pass of the teacher model $\hat{M}$ 
%and student model $M$ respectively.
%Thus for each paired samples $x^S$ and $\tilde{x}^T$ which is generated through CycleGAN and satisfies $\tilde{x}^T = G^{ST}(x^S)$,
%there exist three kinds of feature maps, which are 1) $\hat{A}^S$ obtained by passing source sample $x^S$ through the teacher model $\hat{M}$,
%2) $A^S$ obtained by passing source sample $x^S$ through the student model $M$, 
%3) $A^T$ obtained by passing synthetic target sample $\tilde{x}^T$ through the student model $\hat{M}$.
%We penalize the distance between $\hat{A}^S$ and $A^S$ and that between $\hat{A}^T$ and $A^T$ to align the attention mechanism of the student 
%with the teacher.

Formally, the attention alignment penalty can be formulated as,
\begin{align}
%\mathcal{L}^{AT}{(x_{i^{'}}^S, x_{j^{'}}^T, \{x_j^S, \tilde{x}_j^T\}, \{x_j^T, \tilde{x}_j^S\})}
\mathcal{L}^{AT} &= \sum_{l}\{\sum_{i}{\norm{\frac{A^S_{l}(x_i^S)}{\norm{A^S_{l}(x_i^S)}_{2}}-\frac{A^T_{l}(x_i^S)}{\norm{A^T_{l}(x_i^S)}_{2}}}_{2}}
                 + \sum_{j}{\norm{\frac{A^S_{l}(x_j^S)}{\norm{A^S_{l}(x_j^S)}_{2}}-\frac{A^T_{l}(\tilde{x}_j^T)}{\norm{A^T_{l}(\tilde{x}_j^T)}_{2}}}_{2}} \nonumber \\
                 &+ \sum_{m}{\norm{\frac{A^S_{l}(\tilde{x}_m^S)}{\norm{A^S_{l}(\tilde{x}_m^S)}_{2}}-\frac{A^T_{l}(\tilde{x}_m^S)}{\norm{A^T_{l}(\tilde{x}_m^S)}_{2}}}_{2}} 
                 + \sum_{n}{\norm{\frac{A^S_{l}(\tilde{x}_n^S)}{\norm{A^S_{l}(\tilde{x}_n^S)}_{2}}-\frac{A^T_{l}(x_n^T)}{\norm{A^T_{l}(x_n^T)}_{2}}}_{2}}\},
\label{eq-at-loss}
\end{align}
where the subscript $l$ denotes the layer and ${i}$, ${j}$ denote the samples. The $A^S_l$ and $A^T_l$ represent the attention maps \emph{w.r.t} layer $l$ for the source network and 
the target network, respectively.  $x^S$ and ${x^T}$ are real source and real target domain data, respectively. The synthetic target data ${\tilde{x}_i^T}$ and synthetic source data ${\tilde{x}_n^S}$ satisfy $\tilde{x}_i^T = G^{ST}(x_i^S)$ and $\tilde{x}_n^S = G^{TS}(x_n^T)$, respectively. 

Through Eq. (\ref{eq-at-loss}), the distances of attention maps for the paired images (\emph{i.e.,} (${x_j^S}$, ${\tilde{x}_j^T}$) and
(${x_n^T}$, ${\tilde{x}_n^S}$)) are minimized.
Moreover, we additionally penalize the attention maps of the same input (\emph{i.e.,} ${x_i^S}$ and ${\tilde{x}_m^S}$) passing through different networks.
The attention alignment penalty $\mathcal{L}^{AT}$ allows the attention mechanism to be gradually adapted to the target domain, 
which makes the attention mechanism of the target network invariant to the domain shift.

\textbf{Discussion.}
On minimizing the discrepancy across domains, our method shares similar ideas 
with DAN \cite{long2015learning} and JAN \cite{long2017deep}.
The difference is that our method works on the convolutional layers where the critical structure information 
is captured and aligned across domains;
in comparison, DAN and JAN focus on the FC layers where high-level semantic 
information is considered.
Another notable difference is that 
our method deals with the image-level differences through CycleGAN data pairing,
whereas DAN and JAN consider the discrepancy of feature distributions. 

In DAN and JAN, MMD and JMMD criteria are adopted respectively to measure the
discrepancy of feature distributions across domains.
Technically, MMD and JMMD can also be used as attention discrepancy measures.
%alignment penalties. 
However, as to be shown in the experiment part, MMD and JMMD yield inferior performance to the $L_2$ distance enabled by adversarial data pairing in our method. 
The reason is that 
%As the distance measure, MMD or JMMD is not effective in our framework, 
MMD and JMMD are distribution distance estimators: they map the attention maps to 
the Reproducing Kernel Hilbert Space (RKHS) and lose the structure information. 
So they are not suitable for measuring the attention discrepancy across domains.

\subsection{Training with EM} \label{sec-EM}

To make full use of the available data (labeled and unlabeled), we train the target-domain model with a mixture of real and synthetic data from both source and target domains, as illustrated in Fig. \ref{fig:schematic-diagram}.
For the source and its translated synthetic target domain data, we compute the cross-entropy loss between the predictions and ground-truth labels to back-propagate the gradients through the target network.
The cross-entropy loss for the source and corresponding synthetic target domain data can be formulated as follows,
\begin{align}
\mathcal{L}^{CE} &= - [\sum_i\log{p_{\theta}(y_i^S|x_i^S)} + \sum_j\log{p_{\theta}(y_j^S|\tilde{x}_j^T)}],
\label{eq-ce}
\end{align}
where $y^S \in \{1, 2, \cdots, K\}$ denotes the label for the source sample $x^S$ and the translated synthetic target sample $\tilde{x}^T$.
The probability $p_{\theta}(y|x)$ is represented by the $y$-th output of the target network with parameters $\theta$ given the input image $x$.
%The number of source samples is denote by $N^S$.
$\tilde{x}_j^T = G^{ST}(x_j^S)$.
%Note that if $\tilde{x}^T = G^{ST}(x^S)$, then $\tilde{y}^T = y^T$.

For the unlabeled target data, due to the lack of labels, we employ the EM algorithm to optimize the target network.
The EM algorithm can be split into two alternative steps: the (\textbf{E})xpectation computation step and the expectation (\textbf{M})aximization step.
%We take target data for example to illustrate our algorithm which also applies to the synthetic source data.
The objective is to maximize the log-likelihood of target data samples,
\begin{align}
\sum_i{\log{p_\theta(x^T_i)}},
\label{eq-em-root}
\end{align}
In image classification, our prior is that the target data samples belong to $K$ different categories.
We choose the underlying category ${z_i} \in \{1, 2, \cdots, K\}$ of each sample as the hidden variable,
and the algorithm is depicted as follows (we omit the sample subscript and the target domain superscript for description simplicity).

\textbf{(\textit{i}) The Expectation step.}
We first estimate $p_{\theta_{t-1}}(z|x)$ through,
\begin{equation}
p_{\theta_{t-1}}(z|x) = \frac{p_{\theta_{t-1}}(x | z) p(z)}{\sum_{z}{p_{\theta_{t-1}}(x | z) p(z)}},
\label{eq-em-expectation}
\end{equation}
where the distribution ${p_{\theta_{t-1}}(z | x)}$ is modeled by the target network. $\theta_{t-1}$ is the parameters of the target-domain CNN at last training step ${t-1}$. 
%We use deep CNN $M^T$ to model the distribution ${p_{\theta_{t-1}}(z | x)}$, i.e. the ${K}$ outputs denote the probabilities of categories $x$ belongs to respectively.
We adopt the uniform distributions to depict $p(z)$ (\emph{i.e.,} assuming the occurrence probabilities of all the categories are the same) and $p(x)$ (\emph{i.e.,} assuming all possible image instantiations are distributed uniformly in the manifold of image gallery).
In this manner, $p_{\theta_{t-1}}(z|x) = \alpha p_{\theta_{t-1}}(x | z)$ where $\alpha$ is a constant. %is satisfied.

\textbf{(\textit{ii}) The Maximization step.}
Based on the computed posterior $p_{\theta_{t-1}}(z|x)$, our objective is to update $\theta_{t}$ to improve the lower bound of Eq. (\ref{eq-em-root}),
%\mathcal{L}^{EM}(x) 
\begin{equation}
\sum_{z}p_{\theta_{t-1}}(z|x) \log{p_{\theta_{t}}(x|z)}
\label{eq-em-loss}
\end{equation}
Note that we omit $\sum_{z}p_{\theta_{t-1}}(z|x) \log{p(z)}$ because we assume $p(z)$ subjects to the uniform distribution 
which is irrelevant to ${\theta_{t}}$.
Also, because $p_{\theta}(z|x) = p_{\theta}(x | z)$, Eq. (\ref{eq-em-loss}) is equivalent to, 
\begin{equation}
\sum_{z}p_{\theta_{t-1}}(z|x) \log{p_{\theta_{t}}(z|x)}.
\label{eq-em-loss1}
\end{equation}
% TODO: to determine where to put the following discussions
Moreover, we propose to improve the effectiveness and stability of the above EM steps through three aspects

A) Asynchronous update of $p(z|x)$. We adopt an independent network $M^{post}$ to 
estimate $p(z|x)$ and update $M^{post}$ asynchronously, 
\emph{i.e.,} $M^{post}$ synchronizes its parameters $\theta^{post}$ with the target network every $N$ steps:
%\begin{equation}
$\theta^{post}_t = \theta_{\lfloor t/N \rfloor \times N}$.
%\label{eq-em-asyn}
%\end{equation}
In this manner, we avoid the frequent update of $p(z|x)$ and make the training process much more stable.

B) Filtering the inaccurate estimates. Because the estimate of $p(z|x)$ is not accurate, 
we set a threshold ${p_{t}}$ and discard the samples whose maximum value of $p(z|x)$ over ${z}$ is lower than ${p_{t}}$. 

C) Initializing the learning rate schedule after each update of $M^{post}$.
To accelerate the target network adapting to the new update of the distribution $p(z|x)$, we choose to initialize the 
learning rate schedule after each update of $M^{post}$.

Note that for synthetic source data $\tilde{x}^{S} = G^{TS}(x^T)$, we can also apply the modified EM steps for training.
Because $G^{TS}$ is a definite mapping, we assume $p(z|\tilde{x}^S) = p(z|x^T)$.

To summarize, when using the EM algorithm to update the target network with target data and synthetic source data, 
we first compute the posterior $p(z|x^T)$ through network $M^{post}$ which synchronizes with the target network every $N$ steps.
%Then the cross-entropy loss between $p(z|x^T)$ and $p(x^T|z)$, $p(z|\tilde{x}^T)$ and $p(\tilde{x}^T|z)$ is computed
Then we minimize the loss, 
\begin{align}
\mathcal{L}^{EM} &= -\{\sum_i{\sum_{z_i}{p_{\theta^{post}}(z_i|x_i^T)\log{p_\theta(z_i|x_i^T)}}} 
			     + \sum_j{\sum_{z_j}{p_{\theta^{post}}(z_j|x_j^{T})\log{p_\theta(z_j|\tilde{x}_j^S)}}}\}.
\label{eq-em-asyn}
\end{align}
In our experiment, we show that these modifications yield consistent improvement over the basic EM algorithm.

\subsection{Deep Adversarial Attention Alignment}
Based on the above discussions, our full objective for training the target network can be formulated as,
%{(\{x^S_i, y^S_i\}, \{x^S_j, \tilde{x}^T_j, \tilde{y}^T_j\}, x^T_k)}
\begin{equation}
\min_\theta \mathcal{L}^{full} = \mathcal{L}^{CE} + \mathcal{L}^{EM} + {\beta} \mathcal{L}^{AT} 
\label{eq-full-objective}
\end{equation}
where $\beta$ determines the strength of the attention alignment penalty term $\mathcal{L}^{AT}$.

% concentrate on the attention alignment
\textbf{Discussion.} Our approach mainly consists of two parts: attention alignment and EM training. 
%Both steps are promoted by the use of 
%adversarial data pairing which provides natural image correspondences to perform attention alignment. 
%Adversarial attention alignment plays an important role in the success of our approach.
%%On the one hand, 
%Adversarial data pairing provides natural image correspondences to perform attention alignment,
%while attention alignment built on such image correspondences reduces the system reliance on the quality of the synthetic images (\emph{i.e.,} how close it is to the natural target images) 
%to some extent. In this manner, the negative aspects of the synthetic images can be reduced as well. 
%For example, when the background of generated
%images looks unrealistic, these fake images will be less likely to affect the classification performance 
%because through attention alignment, the model is encouraged to pay more attention to the most discriminative parts of the object and is less likely to be drifted by the irrelevant 
%background.
On the one hand, 
attention alignment is crucial for the success of EM training. 
%On the other hand, the EM training also benefits from the supervised data correspondence which provides labeled images in both domains.  
For EM training, there originally exists no constraint that the estimated hidden variable $Z$ is assigned with the semantic meaning aligned with the ground-truth label,
\emph{i.e.，} there may exist label shift or the data is clustered in an undesirable way.
%Sharing parameters with the network whose training is additionally guided by supervised information can alleviate the above issue.
Training with labeled data (\emph{e.g.} source and synthetic target data) and synchronizing $\theta^{post}$ with $\theta$, the above issue can be alleviated.
In addition, attention alignment further regularizes the training process by encouraging the network to focus on the desirable discriminative information. 

On the other hand, EM benefits attention alignment by providing label distribution estimations 
for target data.
EM approximately guides the attention of target network to fit the target domain statistics,
while attention alignment regularizes the attention of target network to be not far 
from source network.
These two seemingly adversarial counterparts cooperate to make the target network 
acquire the attention mechanism which is invariant to the domain shift.

Note that both parts are promoted by the use of adversarial data pairing 
which provides natural image correspondences to perform attention alignment. 
Thus our method is named ``deep adversarial attention alignment".

%We conduct experiments to prove that they are all necessary ingredients for the adaption from source domain to the target domain.

%EM serves for performing attention alignment during the target network training.
%Adversarial attention alignment stabilizes EM training by encouraging the network 
%to focus on the desirable discriminative information. 

%------------------------------------------------------------------------
\section{Experiment}
\subsection{Setup}
\textbf{Datasets.}
We use the following two UDA datasets for image classification.

1) Digit datasets from \textbf{MNIST} \cite{lecun1998gradient} (60,000 training + 10,000 test images) to \textbf{MNIST-M} \cite{ganin2015unsupervised} (59,001 training + 90,001 test images). % introduction to the dataset
MNIST and MNIST-M are treated as the source domain and target domain, respectively. The images of MNIST-M are created by combining MNIST digits with the patches randomly extracted from color photos of BSDS500 \cite{arbelaez2011contour} as their background.

2) \textbf{Office-31} is a standard benchmark for real-world domain adaptation tasks. It consists of 4,110 images subject to 31 categories. 
This dataset contains three distinct domains, 1)  images which are collected from the Amazon website (\textbf{A}mazon domain), 2) web camera (\textbf{W}ebcam domain), and 3) digital SLR camera (\textbf{D}SLR domain) under different settings, respectively.
The dataset is also imbalanced across domains, with 2,817 images in \textbf{A} domain, 795 images in \textbf{W} domain, and 498 images in \textbf{D} domain.
We evaluate our algorithm for six transfer tasks across these three domains, 
including \textbf{A} ${\rightarrow}$ \textbf{W}, \textbf{D} ${\rightarrow}$ \textbf{W}, \textbf{W} ${\rightarrow}$ \textbf{D}, \textbf{A} ${\rightarrow}$ \textbf{D}, 
\textbf{D} ${\rightarrow}$ \textbf{A}, and \textbf{W} ${\rightarrow}$ \textbf{A}.
%and compare our results with previous unsupervised domain adaptation methods.

\noindent \textbf{Competing methods.}
We compare our method with some representative and state-of-the-art approaches, including 
% introduce the methods we compare with
 RevGrad \cite{ganin2015unsupervised}, JAN \cite{long2017deep}, JAN-A \cite{long2017deep}, DSN \cite{bousmalis2016domain} 
and ADA \cite{haeusser2017associative} which
minimize domain discrepancy on the FC layers of CNN.
We compare with the results of these methods reported in their published papers with identical evaluation setting. 
% a brief explanation about not comparing with other methods
For the task MNIST $\rightarrow$ MNIST-M, 
we also compare with PixelDA \cite{bousmalis2017unsupervised}, a state-of-the-art method on this task.
Both CycleGAN and PixelDA transfer the source style to the target domain without modifying its content heavily.
Therefore, PixelDA is an alternative way to generate paired images across domains and is compatible to our framework.
%However, we want to emphasize that because PixelDA aims to generate more genuine target samples
%with labels, it is compatible to our framework. 
%CycleGAN is not the unique way to generate synthetic target samples with labels.
We emphasize that a model capable of generating more genuine paired images will probably lead to higher accuracy using our method.
The investigation in this direction can be parallel and reaches beyond the scope of this paper.

\subsection{Implementation details}\label{sec:details}
% TODO: the proportion of each part of data, the choice for the EM filtering threshold, and describe the learning rate schedule adopted
% TODO: where do you align the attention for a representative architecture
% for mnist
\textbf{MNIST} $\rightarrow$ \textbf{MNIST-M}
The source network is trained on the MNIST training set. When the source network is trained, it is fixed to guide the training of the target network.
The target and the source network are made up of four convolutional layers, where the first three are for feature extraction and the last one acts as a classifier.
We align the attention between the source and target network for the three convolutional layers.
%We adopt Adam to update our network and the initial learning rate is set to 0.001.
%For a mini-batch input data, we fix the proportions of real source data, synthetic target data, 
%real target data and synthetic source data as $0.35$, $0.15$, $0.35$, and $0.15$, respectively, throughout the experiment.
%For EM training, we set the threshold ${p_t} = 1$ so that the network is learned with all the source and synthetic target data before the first update of $M^{post}$. We then set the threshold ${p_t} = 0.95$ afterwards.

% for office-31
\noindent \textbf{Office-31}
To make a fair comparison with the state-of-the-art domain adaptation methods \cite{long2017deep}, we adopt the ResNet-50 \cite{he2016deep,he2016identity} architecture to perform the adaptation tasks on 
Office-31 and we start from the model pre-trained on ImageNet \cite{deng2009imagenet}. We first fine-tune the model on the source domain data and  fix it. The source model is then used to guide the attention alignment of the target network. The target network starts from the 
fine-tuned model and is gradually trained to adapt to the target domain data.
We penalize the distances of the attention maps \emph{w.r.t} all the convolutional layers except for the first convolutional layer. 

Detailed settings of training are demonstrated in the supplementary material.

\subsection{Evaluation}
% \textbf{The benefit of using synthetic data over source-only transfer. }
% From Table \ref{table:cls-digit} and Table \ref{table:cls-office-1}, it can be seen that 
% most of directly adopting source network to target domain cannot achieve satisfactory classification performance, 
% indicating the domain .
% we observe that training the target network with 
% synthetic target domain data always achieves better performance than that trained with source domain data, 
% which implies that the discrepancy between target domain and synthetic target domain data is much smaller than that between 
% target domain and source domain.

% \textbf{The effectiveness of the deep attention mechanism.}

% \textbf{The effectiveness of EM.}

% \textbf{Comparison of different EM variants.}

\textbf{MNIST}  $\rightarrow$ \textbf{MNIST-M}.
The classification results of transferring MNIST to MNIST-M are presented in Table \ref{table:cls-digit}. We arrive at four observations.
First, our method outperforms a series of representative domain adaptation methods 
(\emph{e.g.}, RevGrad, DSN, ADA) with a large margin, 
all of which minimize the domain discrepancy at the FC layers of neural networks.
Moreover, we achieve competitive accuracy (95.6\%) to 
the state-of-the-art result (98.2\%) reported by PixelDA.
%competitive performance, which outperforms the ADA \cite{haeusser2017associative} by +1.4\% (from 94.2\% to 95.6\%). 
Note that technically, PixelDA is compatible to our method, and can be adopted to improve the accuracy of our model. We will investigate this in the future. 
Second, we observe that 
the accuracy of the source network drops heavily when transferred to the target domain (from 99.3\% on source test set to 45.6\% on target test set), which implies the significant domain shift from MNIST to MNIST-M. Third,
we can see that the distribution of synthetic target data is much closer to real target data than real source data, by observing that training with synthetic target data improves the performance over the source network by about +30\%.
%\textcolor{red}{(as illustrated in Fig. xx)}.
Finally, training with a mixture of source and synthetic target data is beneficial for %the model to 
learning domain invariant features, and improves the adaptation 
performance by +3.5\% over the model trained with synthetic target data only. 

Table \ref{table:cls-digit} demonstrates that our EM training algorithm is an effective way to exploit unlabeled target domain data.
Moreover, imposing the attention alignment penalty ${\mathcal{L}^{AT}}$ always leads to noticeable improvement.
%\vspace{-5mm}
\setlength{\tabcolsep}{6pt}
\begin{table}[ht]
%\floatbox[{\capbeside\thisfloatsetup{capbesideposition={right,top},capbesidewidth=6cm}}]{table}
%{\caption{Classification accuracy (\%) for MNIST ${\rightarrow}$ MNIST-M. ``CNN'' denotes the source and target network (Section \ref{sec:details}).
%% ``S'' and ``T${_f}$'' represent training the target network with source data and synthetic target data, respectively.
%% ``T''  and ``S${_f}$'' mean training the target network with target and synthetic source data respectively.}
%The ``S'' and ``T${_f}$'' represent labeled source data and synthetic target data, respectively.
%The ``T''  and ``S${_f}$'' denote unlabeled target data and synthetic source data, respectively.
%}
%% ``Full'' means training the target network with a mixture of real and synthetic data from both source and target domains.}
%\label{table:cls-digit}}
%\scriptsize
%\scriptsize
\begin{center}
\scalebox{0.7}[0.7]{
\begin{tabular}{ l c  c  c   c   c   c }
\toprule
Method &  Train Data & Accuracy (\%) \\
\midrule
RevGrad \cite{ganin2015unsupervised}          & S+T      & 81.5 \\
DSN \cite{bousmalis2016domain}                & S+T      & 83.2 \\ 
%Tri-training \cite{saito2017asymmetric}       & S+T      & 94.2 \\
ADA \cite{haeusser2017associative}       & S+T      & 85.9 \\
PixelDA \cite{bousmalis2017unsupervised}      & S+T+T${_f}$ & \textbf{98.2} \\
\midrule
Ours (wo $\mathcal{L}^{AT}$)                        &  S+T${_f}$+T+S${_f}$   & 93.5 \\
Ours (w $\mathcal{L}^{AT}$)                           &  S+T${_f}$+T+S${_f}$   & \textbf{95.6} \\
\bottomrule
\end{tabular}
\qquad
\begin{tabular}{ l c  c  c   c   c   c }
\toprule
Method &  Train Data & Accuracy (\%) \\
\midrule
CNN                                           & S          & 45.6 \\
CNN                                               &  T${_f}$                & 75.0 \\
CNN                                               &  S+T${_f}$            & 78.5 \\
CNN + $\mathcal{L}^{AT}$                           &  S+T${_f}$           & 85.7 \\
\midrule
Ours (wo $\mathcal{L}^{AT}$)                        &  S+T${_f}$+T+S${_f}$   & 93.5 \\
Ours (w $\mathcal{L}^{AT}$)                           &  S+T${_f}$+T+S${_f}$   & \textbf{95.6} \\
\bottomrule
\end{tabular}
}
\end{center}
\caption{Classification accuracy (\%) for MNIST ${\rightarrow}$ MNIST-M. ``CNN'' denotes the source and target network (Section \ref{sec:details}).
% ``S'' and ``T${_f}$'' represent training the target network with source data and synthetic target data, respectively.
% ``T''  and ``S${_f}$'' mean training the target network with target and synthetic source data respectively.}
The ``S'' and ``T${_f}$'' represent labeled source data and synthetic target data, respectively.
The ``T''  and ``S${_f}$'' denote unlabeled target data and synthetic source data, respectively}
\label{table:cls-digit}
\end{table}
% TODO
%Exploiting EM algorithm largely improves the network's performance.
%Moreover, additionally adding $\mathcal{L}^{AT}$ significantly improves the performance by 7\%, indicating minimizing the discrepancy of the attention maps
%between source and target domain is quite important.

%\vspace{-5mm}
\noindent \textbf{Office-31}.
The classification results based on  ResNet-50 are shown in Table \ref{table:cls-office-1}. 
With identical evaluation setting, we compare our methods with previous transfer methods and variants of our method. We have three major conclusions.
%Through comparison with identical evaluation setting, 
%we compare our methods with previous transfer methods and variants of our method. 
%we have three major conclusions. 
%\vspace{-15mm}
\setlength{\tabcolsep}{4pt}
\begin{table}[ht]
%\tiny
\begin{center}
\scalebox{0.7}[0.7]{
\begin{tabular}{ l  c c  c   c   c   c   c  c}
\toprule
Method & Train Data & A $\rightarrow$ W & D $\rightarrow$ W & W $\rightarrow$ D & A $\rightarrow$ D & D $\rightarrow$ A & W $\rightarrow$ A & Average \\
\midrule
ResNet-50 & S & 68.4 $\pm$ 0.2 & 96.7 $\pm$ 0.1 & 99.3 $\pm$ 0.1 & 68.9 $\pm$ 0.2 & 62.5 $\pm$ 0.3 & 60.7 $\pm$ 0.3 & 76.1 \\
RevGrad \cite{ganin2015unsupervised} & S+T         & 82.0 $\pm$ 0.4 & 96.9 $\pm$ 0.2 & 99.1 $\pm$ 0.1 & 79.7 $\pm$ 0.4 & 68.2 $\pm$ 0.4 & 67.4 $\pm$ 0.5 & 82.2 \\ 
JAN \cite{long2017deep}   & S+T           & 85.4 $\pm$ 0.3 & 97.4 $\pm$ 0.2 & 99.8 $\pm$ 0.2 & 84.7 $\pm$ 0.3 & 68.6 $\pm$ 0.3 & 70.0 $\pm$ 0.4 & 84.3 \\
JAN-A \cite{long2017deep}   & S+T        & 86.0 $\pm$ 0.4 & 96.7 $\pm$ 0.3 & 99.7 $\pm$ 0.1 & 85.1 $\pm$ 0.4 & 69.2 $\pm$ 0.4 & 70.7 $\pm$ 0.5 & 84.6 \\
\midrule
ResNet-50          & T$_f$   & 81.1   $\pm$ 0.2    & 98.5 $\pm$ 0.2 & 99.8 $\pm$ 0.0 &  83.3 $\pm$ 0.3 & 61.0 $\pm$ 0.2  & 60.2 $\pm$ 0.3 & 80.6 \\ 
%synthetic T + AT        &    $\pm$ 0.2          &  $\pm$ 0.3 & 61.3 $\pm$ 0.2  &  $\pm$ 0.2 \\
ResNet-50 & S+T$_f$                    &  81.9  $\pm$ 0.2          & 98.5 $\pm$ 0.2 & 99.8 $\pm$ 0.0 & 83.7 $\pm$ 0.3 & 66.5 $\pm$ 0.2  & 64.8 $\pm$ 0.3 & 82.5\\ 
Ours (wo ${\mathcal{L}^{AT}}$)   & T$_f$+T     &  86.2 $\pm$ 0.2              &  \textbf{99.3} $\pm$ 0.1 & \textbf{100} $\pm$ 0.0  & 86.5 $\pm$ 0.6 & 69.9 $\pm$ 0.6 & 70.2 $\pm$ 0.2 & 85.4 \\
Ours (w ${\mathcal{L}^{AT}}$)   & T$_f$+T     &  86.8 $\pm$ 0.2              &  \textbf{99.3} $\pm$ 0.1 & \textbf{100} $\pm$ 0.0  & 87.2 $\pm$ 0.5 & 71.7 $\pm$ 0.5 & 71.8 $\pm$ 0.1 & 86.1 \\
Ours (wo ${\mathcal{L}^{AT}}$)  & S+T$_f$+T+S$_f$     & \textbf{87.1} $\pm$ 0.3     &  \textbf{99.3} $\pm$ 0.1 & \textbf{100} $\pm$ 0.0  & 87.1 $\pm$ 0.2& 72.3 $\pm$ 0.2 & 72.2 $\pm$ 0.2 & 86.3 \\ 
Ours (w ${\mathcal{L}^{AT}}$)   & S+T$_f$+T+S$_f$     & 86.8 $\pm$ 0.2              &  \textbf{99.3} $\pm$ 0.1 & \textbf{100} $\pm$ 0.0  & \textbf{88.8} $\pm$ 0.4& \textbf{74.3} $\pm$ 0.2 & \textbf{73.9} $\pm$ 0.2 & \textbf{87.2} \\
\bottomrule
\end{tabular}
}
\end{center}
\caption{Classification accuracy (\%) on the Office-31 dataset based on ResNet-50
%compared with the state-of-the-art methods 
%``S'' and synthetic ``T'' denote training the target network with source and synthetic target data, respectively.
}
\label{table:cls-office-1}
\end{table}
\begin{figure}[H]
\begin{center}
\par
%viewport=0 100 595 550
%\includegraphics[scale=0.30]{frame_0012}
\includegraphics[viewport=100 260 655 590, scale=0.16]{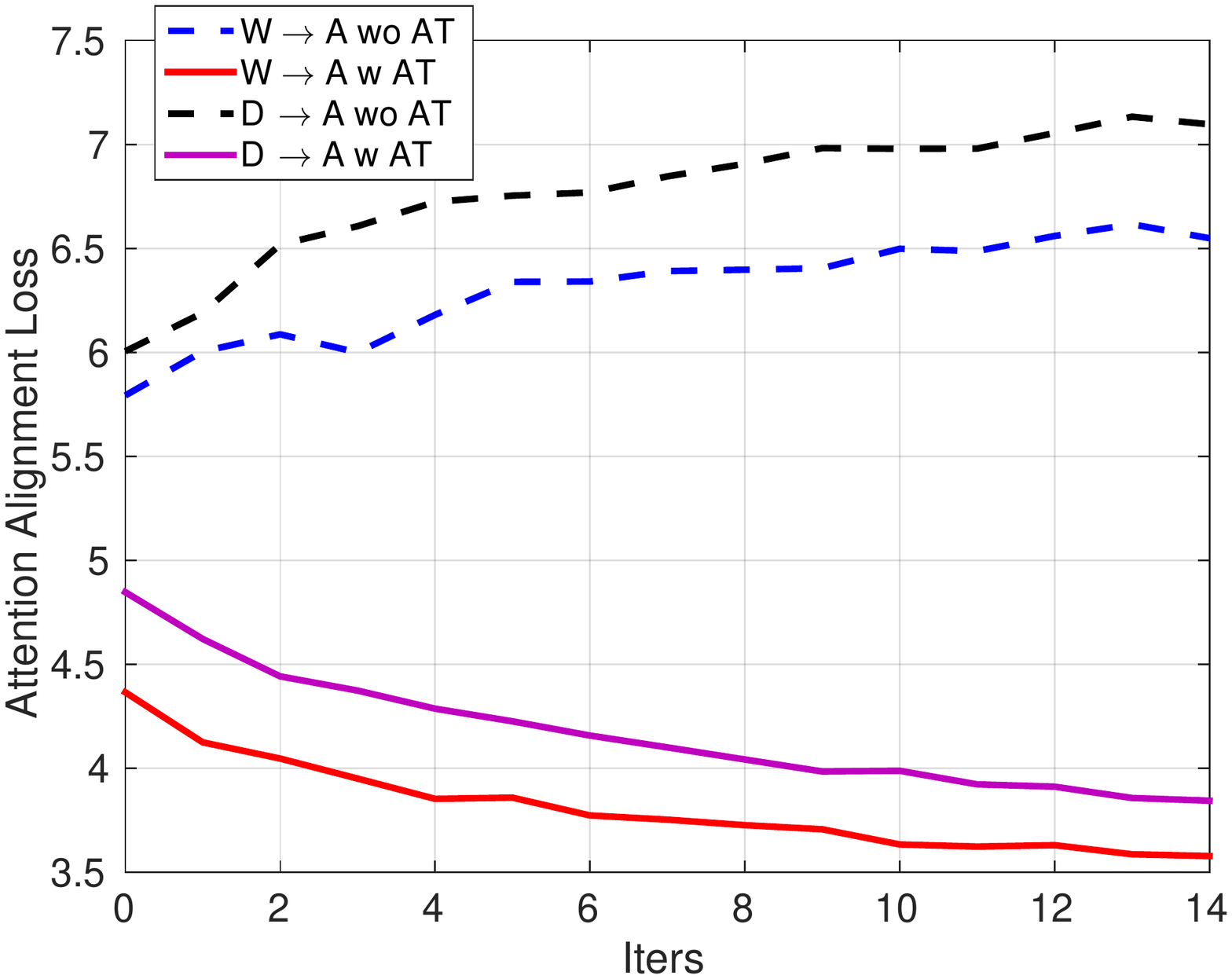}
\qquad
\includegraphics[viewport=50 260 495 590, scale=0.16]{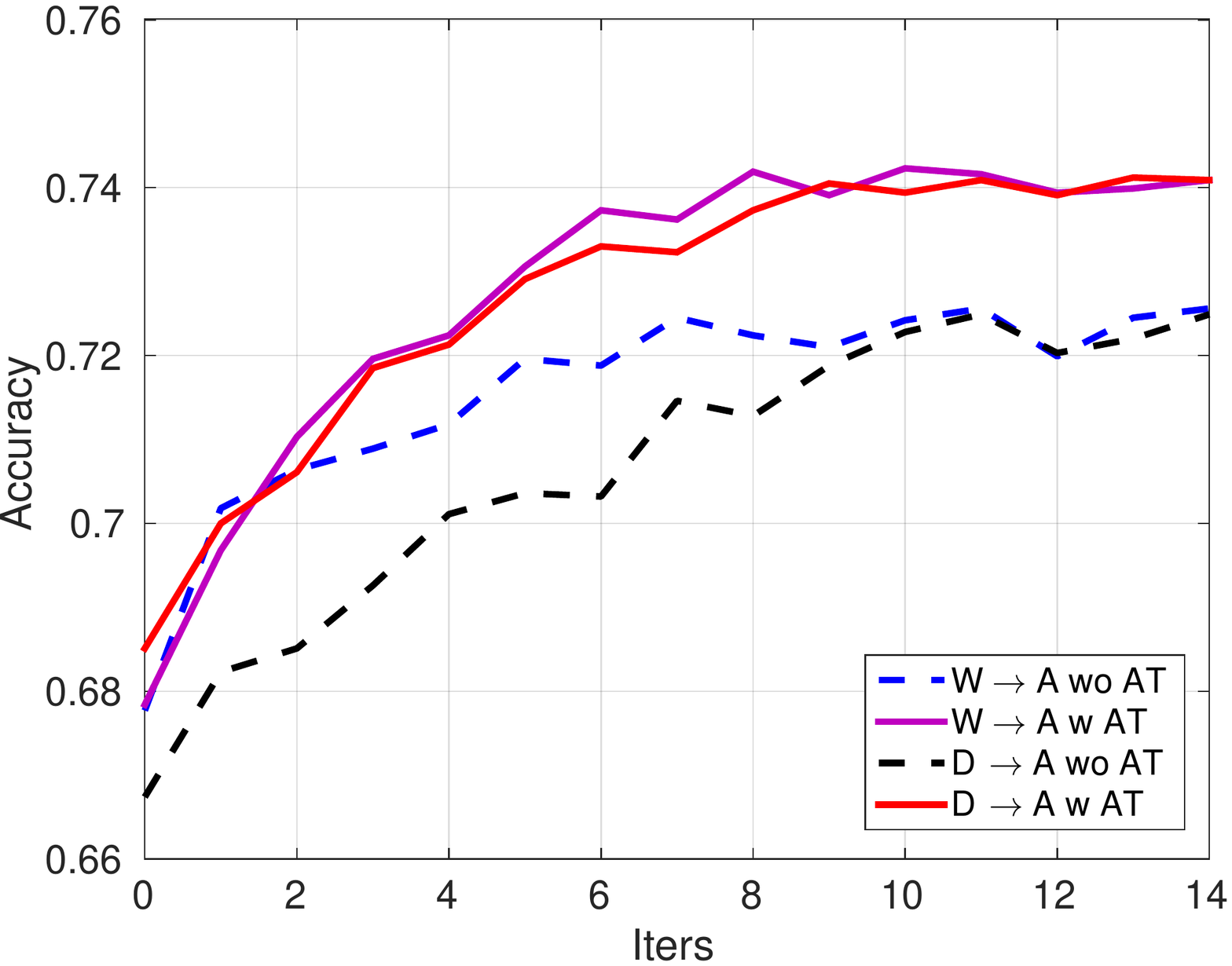}
\end{center}
\caption{\label{fig:test-curve}
{
Analysis of the training process (EM is implemented).
\textbf{Left}:
The trend of ${\mathcal{L}^{AT}}$ during training with and without imposing the ${\mathcal{L}^{AT}}$ penalty term.
\textbf{Right}: The curves of test accuracy on the target domain. 
The results of tasks \textbf{W} ${\rightarrow}$ \textbf{A} and \textbf{D} ${\rightarrow}$ \textbf{A} are presented.
The results for other tasks are similar. 
One iteration here represents one update of the network $M^{post}$ (see Section \ref{sec-EM}).
}
}%\textbf{Left}}
\end{figure}
%\vspace{-3mm}
%\vspace{-5mm}
\setlength{\tabcolsep}{10pt}
\begin{table}%[t]
%\scriptsize
%\tiny
\begin{center}
\scalebox{0.7}[0.7]{
\begin{tabular}{ l  c  c c   c   c  c }
\toprule
Method &  Train Data & A $\rightarrow$ W &  A $\rightarrow$ D & D $\rightarrow$ A & W $\rightarrow$ A & Average \\
\midrule
%&  98.4 $\pm$ 0.1 & 99.8 $\pm$ 0  
%S + synthetic T + $\mathcal{L}^{AT}$        &  81.6  $\pm$ 0.2          & 84.0 $\pm$ 0.3 & 67.2 $\pm$ 0.2  & 65.35 $\pm$ 0.2 \\
%T + EM                    &  87.1  $\pm$ 0.3          & 85.5 $\pm$ 0.3 & 63.2 $\pm$ 0.2  & 70.4 $\pm$ 0.3 \\ 
%T + synthetic S + EM                         &  85.7 $\pm$ 0.4   & 85.3 $\pm$ 0.4 & 60.7 $\pm$ 0.3 &  72.8  $\pm$ 0.1    \\ 
%Ours (T + synthetic S + EM + $\mathcal{L}^{AT}$)    &  86.2 $\pm$ 0.4   & 86.1 $\pm$ 0.2 & 63.5 $\pm$ 0.4  & 71.1  $\pm$ 0.1   \\ 
%T + EM + AT                &  86.0  $\pm$ 0.2          & 86.6 $\pm$ 0.3 & 64.3 $\pm$ 0.2  & 70.4 $\pm$ 0.2 \\
ResNet-50  & S   & 68.4 $\pm$ 0.2 & 68.9 $\pm$ 0.2 & 62.5 $\pm$ 0.3 & 60.7 $\pm$ 0.3  & 65.1 \\
\midrule
EM-A       &  S+T$_f$+T+S$_f$ & 68.6 $\pm$ 0.3     & 73.5 $\pm$ 0.3 & 62.7 $\pm$ 0.3 &  52.8  $\pm$ 0.3  & 64.4 \\ 
EM-A  + $\mathcal{L}^{AT}$ & S+T$_f$+T+S$_f$ & 80.4 $\pm$ 0.2    & 79.1 $\pm$ 0.2 & 66.4 $\pm$ 0.2 &  58.4 $\pm$ 0.2   & 71.1\\ 
%EM-B                       & * $\pm$ 0.3     & * $\pm$ 0.2& * $\pm$ 0.2 & * $\pm$ 0.2   \\ 
%EM-B  + $\mathcal{L}^{AT}$ & * $\pm$ 0.3     & * $\pm$ 0.2&  $\pm$ 0.2 &  $\pm$ 0.2   \\ 
EM-C                   & S+T$_f$+T+S$_f$  & 86.4  $\pm$ 0.3     & 87.0 $\pm$ 0.3 & 69.5 $\pm$ 0.3 & 71.4 $\pm$ 0.3 & 78.6 \\ 
EM-C + $\mathcal{L}^{AT}$  & S+T$_f$+T+S$_f$ & 86.2  $\pm$ 0.2   & 86.6 $\pm$ 0.3 & 71.8 $\pm$ 0.3 & 73.7  $\pm$ 0.2  & 79.6 \\ 
EM-B & S+T$_f$+T+S$_f$ & \textit{very low}   & \textit{very low} & \textit{very low} & \textit{very low} & \textit{very low} \\ 
EM-B + $\mathcal{L}^{AT}$  & S+T$_f$+T+S$_f$ & \textit{very low}   & \textit{very low} & \textit{very low} & \textit{very low} & \textit{very low} \\ 
\midrule
Ours (wo $\mathcal{L}^{AT}$) &  S+T$_f$+T+S$_f$ & \textbf{87.1} $\pm$ 0.3      & 87.1 $\pm$ 0.2 & 72.3 $\pm$ 0.2 & 72.2 $\pm$ 0.2 & 79.7 \\ 
Ours (w $\mathcal{L}^{AT}$) & S+T$_f$+T+S$_f$ & 86.8  $\pm$ 0.2    & \textbf{88.8} $\pm$ 0.4 & \textbf{74.3} $\pm$ 0.2 & \textbf{73.9}  $\pm$ 0.2 & \textbf{80.9} \\
\bottomrule
\end{tabular}
}
\end{center}
\caption{Variants of the EM algorithm with and without $\mathcal{L}^{AT}$. The EM algorithm without asynchronous update of $M^{post}$ is denoted by EM-A, while that without filtering the noisy data %using threshold $p_t$ 
is denoted by EM-B. 
EM-C represents EM training without initializing the learning rate schedule when ${M^{post}}$ is updated 
%See Section \ref{sec-EM} for more detail of our algorithm.
}
\label{table:cls-office-0}
\end{table}
%\vspace{-5mm}
\begin{table}
%\floatbox[{\capbeside\thisfloatsetup{capbesideposition={right,top},capbesidewidth=4cm}}]{table}
%[\FBwidth]
%{\caption{Comparison of the classification accuracy on Office-31 with different attention discrepancy measures.}
%\label{tab:aa-criteria}}
%\scriptsize
%\scriptsize
\begin{center}
\scalebox{0.7}[0.7]{
\begin{tabular}{ l  c c c c  c c}
\toprule
Measure & \textbf{A} $\rightarrow$ \textbf{W} & \textbf{A} $\rightarrow$ \textbf{D} & \textbf{D} $\rightarrow$ \textbf{A} & \textbf{W} $\rightarrow$ \textbf{A} & Average\\
\midrule
$L_1$-norm       & \textit{very low} & \textit{very low} & \textit{very low} & \textit{very low} & \textit{very low}\\
MMD              & 84.7 & 84.1 & 66.2 & 64.5  & 74.9 \\
JMMD             & 85.9 & 85.3 & 70.1 & 71.1  & 78.1 \\
\midrule
Ours             & \textbf{86.8} & \textbf{88.8} & \textbf{74.3} & \textbf{73.9} & \textbf{80.9} \\
\bottomrule
\end{tabular}
}
\caption{Comparison of different attention discrepancy measures on Office-31}
\label{tab:aa-criteria}
\end{center}
%\vspace{-5mm}
\end{table}

First, from Table \ref{table:cls-office-1}, it can be seen that our method outperforms the state of art in all the transfer tasks with a large margin. The improvement is larger  on  harder transfer tasks, where the source domain is substantially different from and has much less data than the target domain,
\emph{e.g.} \textbf{D} $\rightarrow$ \textbf{A}, and \textbf{W} $\rightarrow$ \textbf{A}. 
Specifically, we improve over the state of art result by +2.6\% on average, and by +5.1 \% for the difficult transfer task \textbf{D} $\rightarrow$ \textbf{A}.

Second, we also compare our method with and without the adversarial attention alignment loss ${\mathcal{L}^{AT}}$. 
Although for easy transfer tasks, the performance of these two variants are comparable, 
when moving to much harder tasks,
we observe obvious improvement brought by the adversarial attention alignment, \emph{e.g.,} training with adversarial attention 
alignment outperforms that without attention alignment by $+2\%$ for the task \textbf{D} $\rightarrow$ \textbf{A}, 
and ${+1.7\%}$ for the task \textbf{W} $\rightarrow$ \textbf{A}. 
This implies that adversarial attention alignment helps reduce the discrepancy across domains and regularize the training of the target model.

Third, we validate that augmenting with synthetic target data to facilitate the target network training brings significant improvement of accuracy over source network. This indicates that the discrepancy between synthetic and real target data is much smaller.  
We also notice that in our method, 
the accuracy of the network trained with real and synthetic data from both domains
is much better than 
the one purely trained with real and synthetic target data.
This verifies the knowledge shared by the source domain can be 
sufficiently uncovered by our framework to improve the target network performance. 

Fig. \ref{fig:test-curve} illustrates how the attention alignment penalty $\mathcal{L}^{AT}$ changes during the training process with and without this penalty imposed. 
Without attention alignment, the discrepancy of the attention maps between the source and target network is significantly larger and increases as the training goes on. 
The improvement of accuracy brought by adding $\mathcal{L}^{AT}$ penalty to the objective can be attributed to the much smaller discrepancy of attention maps between the source and the target models, \emph{i.e.,} better aligned attention mechanism.
The testing accuracy curves on the target domain for tasks \textbf{D} $\rightarrow$ \textbf{A} and \textbf{D} $\rightarrow$ \textbf{A} are also drawn in Fig. \ref{fig:test-curve}. It can be seen that the test accuracy steadily increases and 
the model with $\mathcal{L}^{AT}$ converges much faster than that without any attention alignment.

Visualization of the attention maps of our method is provided in Fig. \ref{fig:attention-image}. We observe that through attention alignment, the attention maps of the target network adapt well to the 
target domain images, and are even better than those of the target model trained on labeled target images.

\subsection{Ablation Study}
Table \ref{table:cls-office-0} compares the accuracy of different EM variants. We conduct ablation studies by removing one component from the system at a time (three components are considered which are defined in Section \ref{sec-EM}).
For each variant of EM, we also evaluate the effect of imposing $\mathcal{L}^{AT}$ by comparing training with and without $\mathcal{L}^{AT}$.
By comparing the performances of EM-A, EM-B, EM-C and full method we adopted, we find that the three modifications all contribute considerably to the system. Among them, filtering the noisy data is the most important factor.
We also notice that for EM-A and EM-C, training along with $\mathcal{L}^{AT}$ always leads to a significant improvement, implying performing attention alignment 
is an effective way to improve the adaptation performance.

%\vspace{-5mm}
\subsection{Comparing Different Attention Discrepancy Measures}
%We conduct experiments to evaluate two kinds of criterion to perform attention alignment.
%One represents the way of aggregation for computing attention maps (Table xx),
%and the other measures the discrepancy of attentions across domains (Table xxx).
In this section, we provide a method comparison in measuring the attention discrepancy across domains which is discussed in Section \ref{sec:attention}. 
This paper uses the $L_2$ distance, and the compared methods include the $L_1$ distance, MMD \cite{long2015learning} and JMMD \cite{long2017deep}. Results are presented in Table \ref{tab:aa-criteria}.

We find that 
our method achieves the best results among the four measures.
The $L_1$ distance fails in training a workable network %(with quite low accuracy) 
because it is misled by the noise in the attention maps.
%Considering the mean accuracy over four typical transfer tasks on Office-31 
%(\emph{i.e.} \textbf{A} $\rightarrow$ \textbf{W}, \textbf{A} $\rightarrow$ \textbf{D}, \textbf{D} $\rightarrow$ \textbf{A}, and 
%\textbf{W} $\rightarrow$ \textbf{A}), 
Our method %(with accuracy 80.9\%) 
outperforms MMD/JMMD %(with accuracy 74.9\%) 
%and JMMD (with accuracy 78.1\%) 
by a large margin, 
because our method preserves the structure information, 
as discussed in Section \ref{sec:attention}.

% \subsection{Visualization}
% The aim of attention alignment is to enable the target network to attend to the discriminative parts of an image 
% before making predictions. Fig. \ref{fig:test-curve} has shown that imposing attention alignment penalty effectively reduces 
% the distance of the attention maps between the target network and the source network which is trained on labeled source data 
% and has a well-aligned attention mechanism.

% In Fig. \ref{fig:trained-attention-image}, we visualize the attention maps for the same target image, obtained by passing 
% the image through a source network, a target network trained on labeled target data, and a target network trained by 
% adversarial attention alignment without using any target label information. 
% It demonstrates that through attention alignment, the attention of the target network can be well adapted to the 
% target domain data, and is even better than the one obtained through its own domain labeled data.

\section{Conclusion}
In this paper, we make two contributions to the community of UDA. First, from the \emph{convolutional layers}, we propose to align the attention maps of the source network and target network  
%Through minimizing the discrepancy of attention mechanisms across domains, 
to make the knowledge from source network better adapted to the target one.
Second, from an \emph{EM perspective}, we maximize the likelihood of unlabeled target data, 
%Along with the labeled source data, the EM algorithm 
which enables target network to leverage more training data 
for better domain adaptation. 
Both contributions benefit from the unsupervised image correspondences provided by CycleGAN. 
%use CycleGAN to build the data correspondence across domains and adopt the model trained on the source domain data as a ``teacher" to guide the attention alignment of the ``student" model to be adapted to the target domain. 
%Additionally,  through EM steps. 
Experiment demonstrates that the two contributions both have positive effects on the system performance, and they cooperate together to achieve competitive or even state-of-the-art results on two benchmark datasets.

\textbf{Acknowledgement.}
%Yi Yang is a recipient of the Google Faculty Research Award. 
We acknowledge the Data to Decisions CRC (D2D CRC) and Cooperative Research Centres Programme 
for funding the research.

\bibliographystyle{splncs04}
\bibliography{ata}
\newpage
\renewcommand{\thetable}{A\arabic{table}}   
\renewcommand{\thefigure}{A\arabic{figure}}
\title{Supplementary Material for Paper:\\ 
Deep Adversarial Attention Alignment for Unsupervised Domain Adaptation: \\
the Benefit of Target Expectation Maximization} 
% Replace with your title

\titlerunning{Supplementary Material}
% Replace with a meaningful short version of your title
%
\author{Guoliang Kang\inst{1} \and
Liang Zheng\inst{1,2} \and
Yan Yan\inst{1} \and 
%Zikun Liu\inst{3} \and 
Yi Yang\inst{1}}
%
%Please write out author names in full in the paper, i.e. full given and family names. 
%If any authors have names that can be parsed into FirstName LastName in multiple ways, please include the correct parsing, in a comment to the volume editors:
%\index{Lastnames, Firstnames}
%(Do not uncomment it, because you may introduce extra index items if you do that, we will use scripts for introducing index entries...)
\authorrunning{Guoliang Kang \emph{et al.}}
% Replace with shorter version of the author list. If there are more authors than fits a line, please use A. Author et al.
%

\institute{CAI, University of Technology Sydney \\
\email{\{Guoliang.Kang@student., Yan.Yan-3@student., Yi.Yang@\}uts.edu.au} \and 
Research School of Computer Science, Australian National University \\
\email{liangzheng06@gmail.com}
%\and
%Samsung Research Institute China - Beijing(SRC-B) \\
%\email{zikun.liu@samsung.com}
}
\maketitle              % typeset the header of the contribution
\section{Training Details}
% TODO: the proportion of each part of data, the choice for the EM filtering threshold, and describe the learning rate schedule adopted
% TODO: where do you align the attention for a representative architecture
% for mnist
\textbf{MNIST} $\rightarrow$ \textbf{MNIST-M}
We adopt Adam to update our network and the initial learning rate is set to 0.001.
For a mini-batch input data, we fix the proportions of real source data, synthetic target data, 
real target data and synthetic source data as $0.35$, $0.15$, $0.35$, and $0.15$, respectively, throughout the experiment.
For EM training, we set the threshold ${p_t} = 1$ so that the network is learned with all the source and synthetic target data before the first update of $M^{post}$. We then set the threshold ${p_t} = 0.95$ afterwards.

\noindent \textbf{Office-31}
We follow the same learning rate schedule adopted in \cite{long2017deep} throughout our experiment
except that we initialize the learning rate schedule after each update of posterior estimation network $M_{post}$ (see Section 3.3 of the text).
For a mini-batch input data, we fix the proportions of real source data, synthetic target data, 
real target data and synthetic source data as $0.35$, $0.15$, $0.35$, and $0.15$ respectively, throughout our experiment.
Threshold ${p_t}$ for EM training is set as ${0.95}$.
We choose ${\beta}$ through validation following the same protocol as \cite{long2017deep}.
%\emph{e.g.}. 
%for task \textbf{W} $\rightarrow$ \textbf{A}, 
%we use the dataset \textbf{D} as the validation set.

%\section{Additional Note}
In the experiment of Office-31, we do not penalize the distances between attention maps 
\emph{w.r.t} the first convolutional layer and the max-pooling layers of ResNet-50, because
1) Attention of the first convolutional layer focuses on low-level details and is easily affected by noise. 
2) The max-pooling layer does not have parameters (totally determined by the outputs of previous convolutional layer). 
So it is not necessary to additionally align its attention.
3) We empirically find that ignoring these layers when performing attention alignment brings no loss of accuracy 
but is more efficient in computation.

\section{Impact of Hyper-parameters}
\begin{figure}[t]
\begin{center}
\par
%viewport=0 100 595 550
%\includegraphics[scale=0.30]{frame_0012}
\includegraphics[viewport=100 300 495 580, scale=0.30]{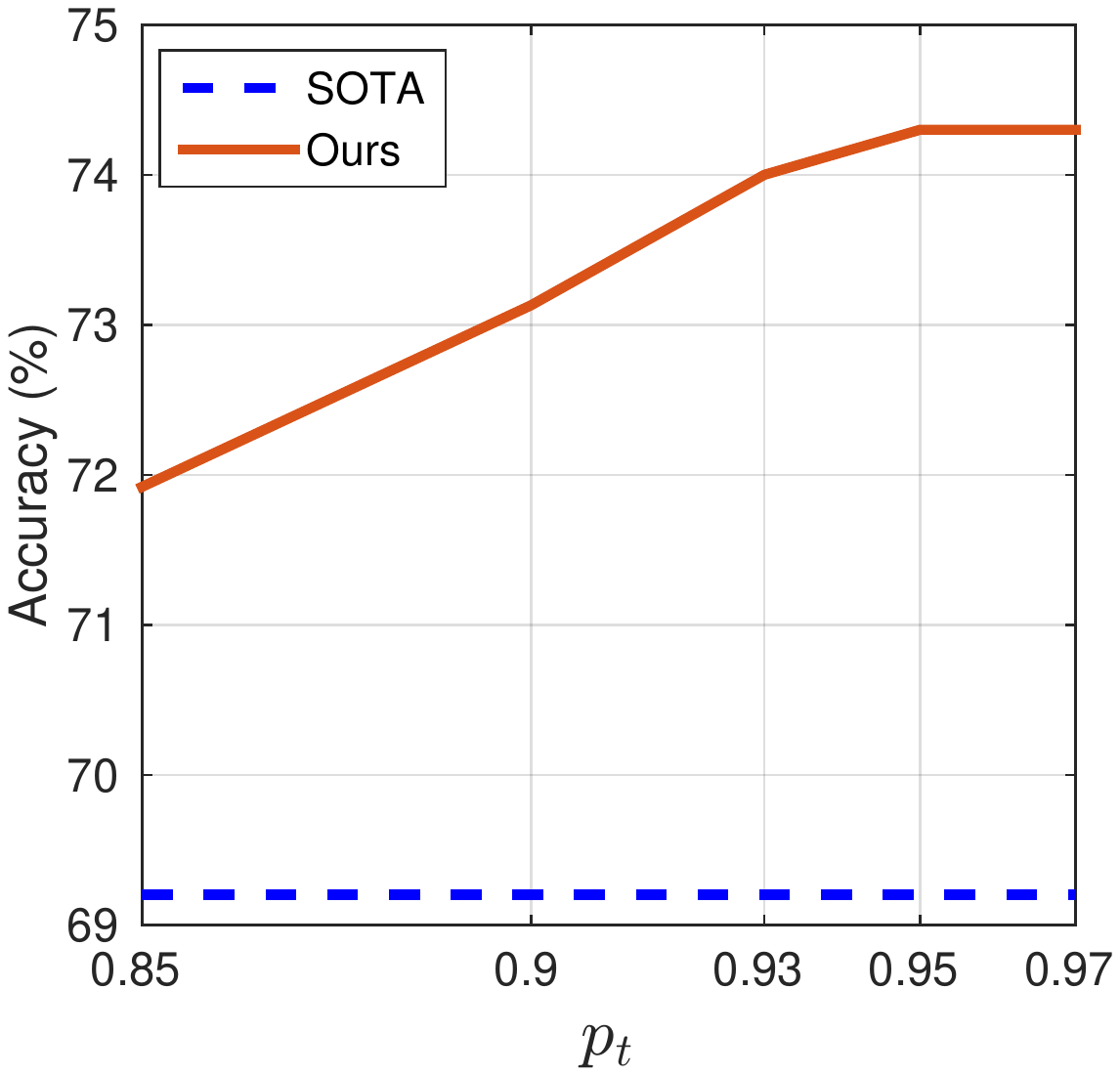}
\includegraphics[viewport=130 300 495 580, scale=0.30]{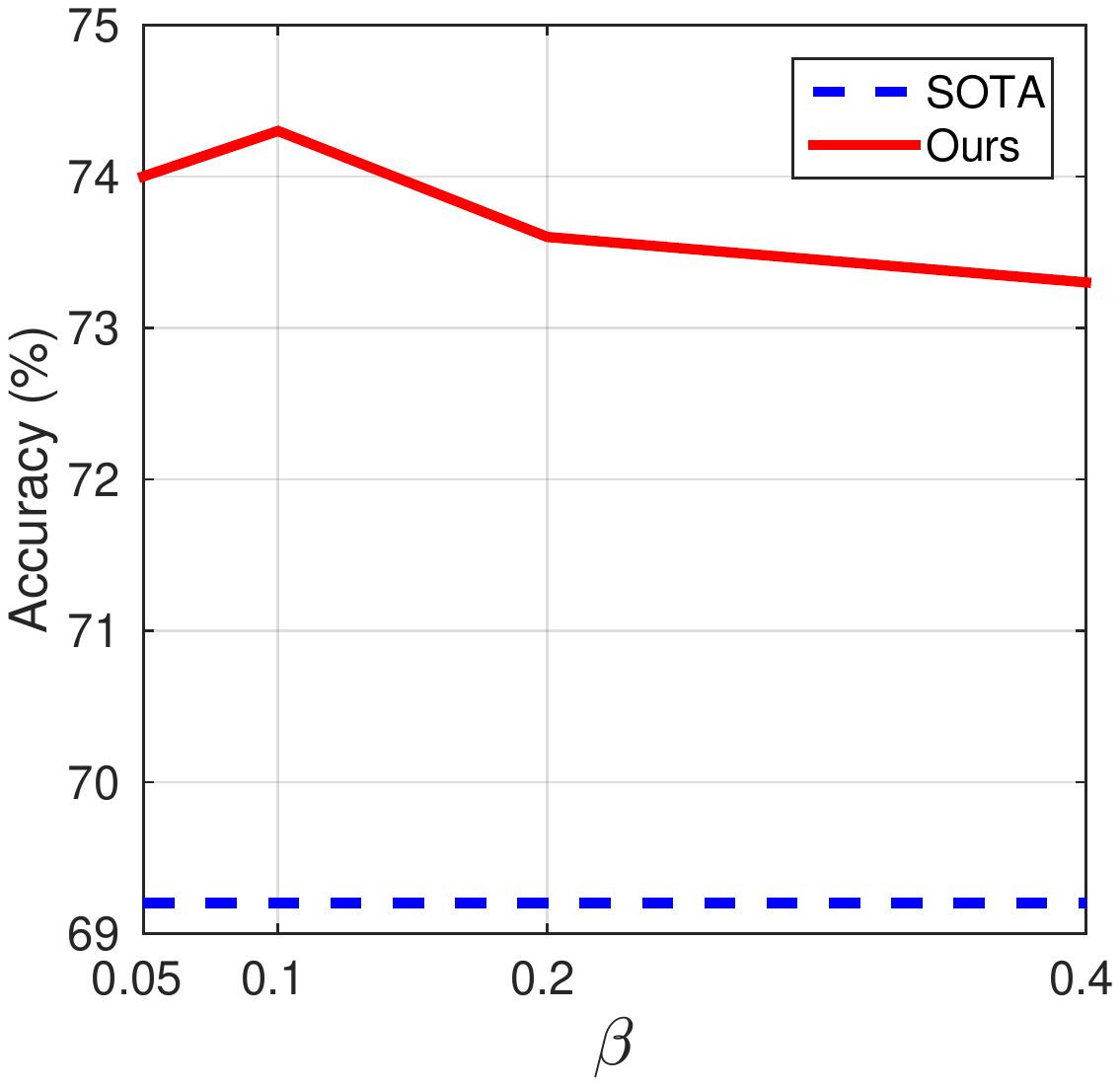}
\end{center}
\caption{\label{fig:curve}
{
The impact of hyper-parameters on the classification accuracy of target model. 
The results for task \textbf{D} $\rightarrow$ \textbf{A} on Office-31 are illustrated, 
with a comparison to the previous state-of-the-art (SOTA). 
The trends are similar for other tasks.
\textbf{Left}:
Accuracy vs. ${p_t}$.
\textbf{Right}: 
Accuracy vs. ${\beta}$.
}
}
\end{figure}

We investigate the impact of $p_t$ (\emph{i.e.} filtering threshold in EM) and $\beta$ (\emph{i.e.} the strength of attention alignment penalty) 
on the classification accuracy of target model, respectively. 
The results are shown in Fig. \ref{fig:curve}.

Within a range, a larger $p_t$ leads to better accuracy,
while with the growth of $\beta$, the accuracy of the model increases before the slightly decrease.
For both $p_t$ and $\beta$, we observe that within a wide range, the accuracy of our method outperforms the previous state-of-the-art method
with a large margin, which implies the superiority of our method.

\begin{table}[t]
%\floatbox[{\capbeside\thisfloatsetup{capbesideposition={right,top},capbesidewidth=4cm}}]{table}
%[\FBwidth]
%{\caption{Comparison of the classification accuracy on Office-31 with different attention discrepancy measures.}
%\label{tab:aa-criteria}}
%\scriptsize
\scriptsize
\begin{center}
\begin{tabular}{ l  c c c c  c c}
\toprule
Variant & \textbf{A} $\rightarrow$ \textbf{W} & \textbf{A} $\rightarrow$ \textbf{D} & \textbf{D} $\rightarrow$ \textbf{A} & \textbf{W} $\rightarrow$ \textbf{A} & Average\\
\midrule
$L_1$                     & 85.2 & 87.8 & 73.3 & 73.0 & 79.8 \\
$L_{\infty}$              & 86.4 & 87.2 & 73.1 & 73.2  &  80.0 \\
FM                        & 86.0 & 87.6 & 73.2 & 72.9  &  79.9 \\
\midrule
Ours (attention)             & \textbf{86.8} & \textbf{88.8} & \textbf{74.3} & \textbf{73.9} & \textbf{80.9} \\
\bottomrule
\end{tabular}
\caption{Comparison of aligning different representations on Office-31}
\label{tab:aa-variants}
\end{center}
\end{table}

\section{Comparison with different variants of attention}
We conduct experiment to verify the effectiveness of attention defined by Eq. (4) of the text.
The comparison results are summarized in Table \ref{tab:aa-variants}.
Note that the attention mechanism defined in our method is 
the aggregation of feature maps along channels using $L_2$-norm, and $L_1$- and $L_\infty$-norm aggregating methods are compared in Table \ref{tab:aa-variants}. 
We also compare our method with directly aligning feature maps without any aggregation (denoted as ``FM'' in Table \ref{tab:aa-variants}).
We find that aligning the proposed attention performs much better than aligning other variants, 
which verifies the effectiveness of attention defined by Eq. (4).

\end{document}